# Developing an Optimal Model for Predicting the Severity of Wheat Stem Rust
# (Case study of Arsi and Bale Zone)

Tewodrose Altaye

Department of Information Technology, Arba Minch University

, Arba minch, Ethiopia

tewodrosealtaye@gmail.com

## Abstract

This research utilized three types of artificial neural network (ANN) methodologies, namely Backpropagation Neural Network (BPNN) with varied training, transfer, divide, and learning functions; Radial Basis Function Neural Network (RBFNN); and General Regression Neural Network (GRNN), to forecast the severity of stem rust. It considered parameters such as mean maximum temperature, mean minimum temperature, mean rainfall, mean average temperature, mean relative humidity, and different wheat varieties. The statistical analysis revealed that GRNN demonstrated effective predictive capability and required less training time compared to the other models. Additionally, the results indicated that total seasonal rainfall positively influenced the development of wheat stem rust.

Keywords: Wheat stem rust, Back propagation neural network, Radial Basis Function Neural Network, General Regression Neural Network.

1. **Introduction**

   Wheat is among the major cereal crops cultivated in Ethiopia. Ethiopia is the second largest producer of wheat in sub-Saharan Africa [1]. The crop has considered as the main staple food of Ethiopian population particularly in highlands of the country [2],where it has produced in a large volume and 95% of the total production is produced by small scale farmers. Wheat ranks first in area coverage (23.78%) and second in total production (3.53%) in Oromia Regional State. Arsi and Bale highlands are the major wheat producing zones of Ethiopia are deemed to be the wheat belts of East Africa [3].

   According to [4] , The production and productivity of wheat in Ethiopia has increased in the last decades, though the national average yield has not exceeded 21.25 tons/ha. This is by far below the world's average yield/ha which is about 33.3 tones/ha [5]. This low yield is attributed to multi-faced abiotic and biotic factors such as cultivation of unimproved low yielding varieties, low and uneven distribution of rainfall, poor agronomic practices, insect pests and serious disease like rusts [6].
   Rust fungal pathogens are among the major stresses that cause high yield losses in wheat crop. Over 30 fungal wheat diseases are identified in Ethiopia, stem rust caused by Puccinia graminis f.sp. tritici is one of the major production constraints in most wheat growing areas of the country; causing yield losses of up to 100% during epidemic years [7]. Bale and Arsi zones are conducive for stem rust epidemics.

   A developing country like Ethiopia uses the traditional way to prevent diseases is to apply fungicides on fixed calendar dates. This method may allow the number of sprays to increase, increase the cost of crop production and the environmental impact of pesticides. Similarly, the government take actions after the disease control cultivated area. As a result of this, the disease capable of causing 100% crop loss within weeks [9]. The early warning system is the best method to control the prevalence of wheat stem rust.

   The early warning system is the best method to control the prevalence of wheat stem rust. Forecast the future severity and production loss of wheat, this enables them to prevent the recurrent occurrence of the disease. An example from England illustrates this clearly and there are probably many tropical plant diseases where the same situation applies. An example from England illustrates this clearly and there are probably many tropical plant diseases where the same situation applies [10].
   In rigorous analysis, there are numerous efforts have been made the importance of prediction on the severity of wheat rust disease using both Statistical and machine learning method.
   Wheat stem rust is influenced by various factors, including pathogen race, variety resistance and meteorological conditions and others. The relationship between wheat stem rust and these factors is usually nonlinear. Neural networks have good advantages of solving the nonlinear problems.

Neural networks have good advantages of solving the nonlinear problems and they can find the most effective factors on wheat stem rust disease. They are composed of interconnecting artificial neurons. They have good adaptive learning ability and nonlinear mapping ability.

Neural networks have been successfully applied in various fields, including pattern recognition, image analysis, and control systems. Vast literature about ANNs, basically in the empirical field, showed that ANNs comparability or superiority to conventional methods for estimating functions [15] [16].

This study developed a predicting model for wheat stem rust severity using Artificial Neural networks with highly accurate and effective way.

## 2. Method and Methodologies

### 2.1 Description of study area

The study was conducted at two locations, namely in the selected district in Arsi and Bale zone, in the Southeastern Highland of Ethiopia. The geographical maps of study areas are presented here under (figure). Both study areas located in the Oromia regional state and they are representative sites for wheat production for commercial purposes. Bale zone located at 6.7606° N, 40.3089° E and the altitude exceeds 2300 meters above sea level. The Bale zone characterized by bimodal rainfall forming two wheat-growing seasons in a year. The main season locally called Bonaa (extends from July to December) and the other season called Ganna (extends March to June). The mean Annual range is 875 mm. The annual average temperature is 250c. The average annual maximum temperature is 250 c and the average annual minimum temperature 100c. The soil type is dominated by Pellic Vertisols and slightly acidic.

### 2.2 Model Development

In ANN modeling, the choice of the ANN training and architectural parameters is the most vital criteria that determine the degree of success of an ANN model. The selection of these parameters is basically problem dependent. However, there are some guides that help in ANN model development. Generally, in order to develop optimal ANN model of high performance the following modeling steps must be considered carefully: 1) collection and preparation of data for training, validation and testing the ANN models, 2) selection of ANN training and architectural parameters, 3) ANN models training, and 4) testing the ANN models and analysis of the results. The overall design issues are represent diagrammatically as shown figure.

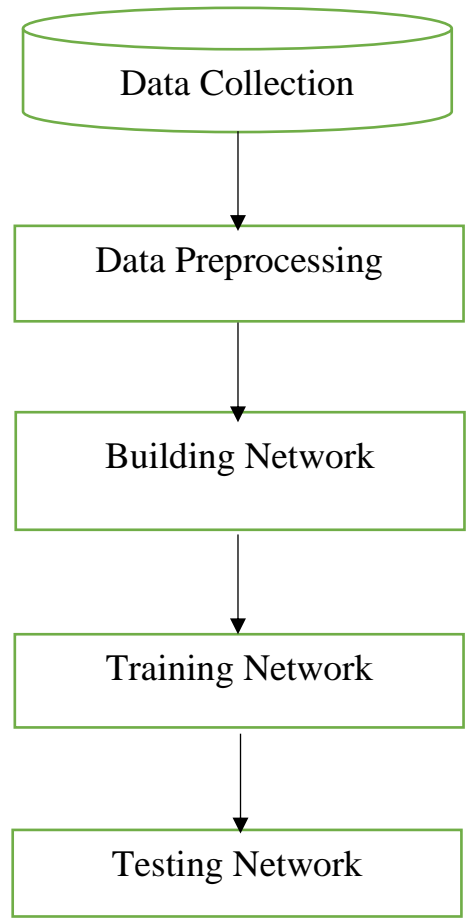

Figure 1.Basic flow for Designing ANN model

**2.3 Model Validation and Testing**

ANN is empirical models; therefore, validation phase is critical to successful training and operation. The purpose of the network validation is to ensure its ability to generalize within limits set by the validation data in a robust fashion, rather than simply memorized the input-output relationships that are contained in the training data.

The coefficients of determination ($R^2$), mean square error (MSE), and the root mean square error (RMSE) are the key criteria that are used to evaluate the performance of ANN model. In general, RMSE and MAE indicate the residual errors, which give a global idea of the difference between the observed and predicted values. $R^2$ is the proportion of variability (sum of squares) in a data set that is accounted for by a model. When the RMSE and MAE are at the minimum and $R^2$ is high ($R^2 > 0.80$), a model can be judged as very good [17]

3. Experiments the Proposed Study models

    3.1 Designing ANN Models

Designing ANN models follows several systematic procedures. In general, there are five basic steps: (1) collecting data, (2) Preprocessing Data (3) building Network, (4) Train, (5) Test Performance of model

### 3.1.1 Data sets and Data Sources

The data set consists of input factors/variables and an output variable. The input factors/variables represent the factor that influences the wheat stem rust. The input variables such as Rainfall, temperature, relative humidity, wheat variety of crop. The output variable is the severity level of wheat stem rust. The data collection phase was carefully planned to ensure the adequacy of data, the basic principles at work are captured and that the data is noise-free.

### 3.1.1.1 Meteorological Data sources

The meteorological data for the year 2000-2018 of selected Aris and Bale zone were obtained from the respective weather stations, which are close to the wheat experimental areas. In the Aris and Bale zones the month of July, August, September, October, November and December as the main seasons and Cropping seasons. Thus, the implemented study explores the data of these month. The input of weather variables is mean monthly rainfall (mm), mean minimum temperature, mean maximum temperature, and average monthly temperature (°C) reckoned from minimum and maximum temperatures and average relative humidity for six months of data were used for the study. The data between 2000 and 2016 is used for training MLP, RBFNN and GRNN based ANN techniques while the data between 2017 and 2018 is used for testing models.

### 3.1.1.2 Wheat disease Data Sources

Wheat stem rust severity data of forty-three commercial wheat cultivars during 2000-2018 was obtained from Ethiopia, Sinana and Kulumsa Agricultural Research Centers. The stem rust data were recorded by estimating the percentage of leaf area affected (Modified Cobb Scale) [18] and infection types were recorded periodically at various stages of plant growth and converted to the coefficient of infection (C.I) using the method employed by Loegering [19]. The time and frequency of observations varied from year to year and from site to site, but the present study used the data recorded at the early dough stage. Rust severity was assessed from whole plots of each treatment. The sources of inoculum were from naturally occurring stem rust spores. Also, the data between 2000 and 2016 is used for training MLP, RBFNN and GRNN based ANN techniques while the data between 2017 and 2018 is used for testing models.

### 3.1.2 Data Pre-processing

The performance of artificial neural networks is highly influenced by the size of the dataset and the data preprocessing. So, after data collection, the data processing procedures are conducted to train the ANNs more efficiently. In order not to saturate the condition of the neurons, data normalization is required. If neurons get saturated, then the changes in the input value will produce a very small change or not change at all in the output value. For this reason, data must be normalized before being presented to the artificial neural network.

In this study, the researcher used the min-max method to normalize and compresses the range of the training data between -1 and 1. Min-Max technique was the best one preserving the overall accuracy compare than other technique [20]. A neural network in MATLAB understand in Matrix form, input and the target data sets coded into numerical value. This study used six parameters to develop an Artificial Neural Network model.

In addition to this, the wheat varieties attribute is string variable, it must be encoded to numerical value to provide in MATLAB Environment. In the output data, the severity level of wheat stem rust was divided into high, medium and low according to [21]and converted into numerical values. So, the output coding system divided into three but one neuron only active at a time. If the severity level high, the first digit only will be active (1,0,0), the severity level medium the (0,1,0) and the severity level low, the third digit only active (0,0,1).

### 3.1.3 Building Network

At this stage, the researcher specifies the number of hidden layers, neuron each layer, transfer function, weight/bias learning function, performance function. In this study, Multilayer Perceptron Neural Network (MLP), Radial Basis Neural Network (RBFNN) and General Regression Neural Network (GRNN) are used.

### 3.1.3.1 Determination of Appropriate Architecture of MLP

I. Input layer

The Input layer is a layer which communicates with the external environment that presents a pattern to the neural network. The input layer should represent the condition for which we are training the neural network. Every input neuron should represent some independent variable that has an influence over the output of the neural network.

Table 1.Summery of input variables

| Design Parameter | Input variables | Numerical code |
|---|---|---|
| X1 | Rainfall | Numerical value |
| X2 | Maximum Temperatures | Numerical value |
| X3 | Minimum Temperatures | Numerical value |
| X4 | Average Temperatures | Numerical Value |
| X5 | Relative Humidity | Numerical value |
| X6 | Wheat Variety | Categorical Value |

II. Output layer

The output layer of the neural network is what actually presents a pattern to the external environment. The pattern presented by the output layer can be directly traced back to the input layer. The severity of wheat stem rust level as output variable. In the output data, the severity level of wheat stem rust was divided into three classes represented by 1,2 and 3 respectively. The three classes normalized and the class 1,2,3 was expressed as (0,0,1), (0,1,0) and (1,0,0) respectively.

Table 2. Summary of Output Variable

| Output Variables | Categorical value | Numerical code | Normalized output value |
|---|---|---|---|
| The severity level of stem rust | High | 3 | (1,0,0) |
| | Medium | 2 | (0.1,0) |
| | Low | 1 | (0,0,1) |

III. Number of Hidden layers

A hidden layer in an artificial neural network is a layer in between input layers and output layers, where artificial neurons take in a set of weighted inputs and produce an output through an activation function. In this study, the researcher used one hidden layer. Because, one hidden layer is sufficient for any universal functional approximation [22] [23].

IV. Number of Hidden neurons

There is no clear-cut method to determining the number of hidden neurons. The most common way in determining the number of hidden nodes is via experiments or by trial-and-error. Several rules of thumb have also been proposed. To help avoid the overfitting problem, some researchers have provided empirical rules to restrict the number of hidden nodes. Lachtermacher and Fuller [24] give a heuristic constraint on the number of hidden nodes. In the case of the popular one hidden layer n0etworks, several practical guidelines exist. These include using "2n +1 [25]", "2n [26], "n" [27] , "n/2" [28], where n is the number of input nodes. However, one of these heuristic choices works well for all problems.

4. **Modeled and Experiment of Proposed MLP Approach**

The goal of this study to predict the severity of wheat stem severity level using Artificial Neural Network and executed in MATLAB 2019a. The architecture of a proposed network was modeled with two layers, one hidden layer and one output layer. The experiments divided organized into five categories to choose the best optimal MLP ANN model to Predict the severity of Wheat stem rust.

I. Experiment one: Selection of Hidden Neurons

In experiment one, the researcher tested hidden neurons from 3-13 tested and was going to select the best hidden neurons using model evaluation methods. Also, the researcher used default training, transfer, divide, learning function to select hidden neurons.

Table 3. The best validation performance of the selected number of hidden neurons

| No | Number of Hidden Neuron | Training function | Transfer function | | Divine function | Learning Function | Best validation performance (MSE) | Epoch |
|---|---|---|---|---|---|---|---|---|
| | | | Hidden layer | Output layer | | | | |
| 1 | 3 | TRAINLM | TANSIG | TANSIG | DIVIDERAND | LERANGDM | 0.061806 | 11 |
| 2 | 4 | TRAINLM | TANSIG | TANSIG | DIVIDERAND | LERANGDM | 0.059907 | 9 |
| 3 | 5 | TRAINLM | TANSIG | TANSIG | DIVIDERAND | LERANGDM | 0.061895 | 8 |

| 4  | 6  | TRAINLM | TANSIG | TANSIG | DIVIDERAND | LERANGDM | **0.056504** | 14 |
| 5  | 7  | TRAINLM | TANSIG | TANSIG | DIVIDERAND | LERANGDM | **0.056365** | 3  |
| 6  | 8  | TRAINLM | TANSIG | TANSIG | DIVIDERAND | LERANGDM | **0.050125** | 19 |
| 7  | 9  | TRAINLM | TANSIG | TANSIG | DIVIDERAND | LERANGDM | 0.065986     | 3  |
| 8  | 10 | TRAINLM | TANSIG | TANSIG | DIVIDERAND | LERANGDM | **0.051989** | 3  |
| 9  | 11 | TRAINLM | TANSIG | TANSIG | DIVIDERAND | LERANGDM | 0.05665      | 5  |
| 10 | 12 | TRAINLM | TANSIG | TANSIG | DIVIDERAND | LERANGDM | **0.054065** | 7  |
| 11 | 13 | TRAINLM | TANSIG | TANSIG | DIVIDERAND | LERANGDM | 0.061069     | 14 |

II. Select divide function

Table 4. Best validation performance results with Divide function

| No | Number of Hidden Neuron | Training function | Transfer function | | Divine function | Learning Function | Best validation performance (MSE) | Epoch |
|----|-------------------------|-------------------|-------------------|---|-----------------|-------------------|-----------------------------------|-------|
|    |                         |                   | Hidden layer | Output layer |          |          |              |    |
| 1  | 6  | TRAINLM | TANSIG | TANSIG | DIVIDEIND | LERANGDM | **0.053712** | 9  |
| 2  | 7  | TRAINLM | TANSIG | TANSIG | DIVIDEIND | LERANGDM | 0.064297     | 4  |
| 3  | 8  | TRAINLM | TANSIG | TANSIG | DIVIDEIND | LERANGDM | 0.064087     | 11 |
| 4  | 10 | TRAINLM | TANSIG | TANSIG | DIVIDEIND | LERANGDM | 0.061059     | 8  |
| 5  | 12 | TRAINLM | TANSIG | TANSIG | DIVIDEIND | LERANGDM | 0.061676     | 6  |

III. Select transfer function

Table 5. The comparison between the transfer functions

| No | Number of Hidden Neuron | Training function | Transfer function | | Divine function | Learning Function | Best validation performance (MSE) | Epoch |
|----|-------------------------|-------------------|-------------------|---|-----------------|-------------------|-----------------------------------|-------|
|    |                         |                   | Hidden layer | Output layer |          |          |              |    |
| 1 | 8 | TRAINLM | LOGSIG | PURELIN | DIVIDERAND | LERANGDM | **0.049962** | 7  |
| 2 | 8 | TRAINLM | TANSIG | TANSIG  | DIVIDERAND | LERANGDM | 0.050125     | 5  |
| 3 | 7 | TRAINLM | LOGSIG | TANSIG  | DIVIDERAND | LERANGDM | 0.050277     | 11 |

IV. Select learning function

Table 6. The comparison between the learning functions

| No | Number of Hidden Neuron | Training function | Transfer function | | Divine function | Learning Function | Best validation performance (MSE) | Epoch |
|---|---|---|---|---|---|---|---|---|
| | | | Hidden layer | Output layer | | | | |
| 3 | 8 | TRAINLM | LOGSIG | PURELIN | DIVIDERAND | LERANGDM | **0.049962** | 7 |
| 6 | 6 | TRAINLM | LOGSIG | PURELIN | DIVIDERAND | LERANGD | 0.053118 | 10 |

V. Selection of training algorithm

Table 7. The comparison between of All training Algorithm

| No | Number of Hidden Neuron | Training function | Transfer function | | Divine function | Learning Function | Best validation performance (MSE) |
|---|---|---|---|---|---|---|---|
| | | | Hidden layer | Output layer | | | |
| 1 | 8 | Trainlm | LOGSIG | PURELIN | DIVIDERAND | LERANGDM | **0.049962** |
| 2 | 7 | Trainbfg | LOGSIG | PURELIN | DIVIDERAND | LERANGDM | 0.058244 |
| 3 | 8 | Trainrp | LOGSIG | PURELIN | DIVIDERAND | DIVIDERAND | 0.054246 |
| 4 | 12 | Traingdx | LOGSIG | PURELIN | DIVIDERAND | LERANGDM | 0.053174 |
| 5 | 8 | Trainscg | LOGSIG | PURELIN | DIVIDERAND | LERANGDM | 0.052486 |
| 6 | 8 | Traincgb | LOGSIG | PURELIN | DIVIDERAND | LERANGDM | 0.054275 |
| 7 | 7 | Trainoss | LOGSIG | PURELIN | DIVIDERAND | LERANGDM | 0.053312 |
| 8 | 8 | Traincgf | LOGSIG | PURELIN | DIVIDERAND | LERANGDM | 0.050915 |
| 9 | 10 | Traingdm | LOGSIG | PURELIN | DIVIDERAND | LERANGDM | 0.083611 |
| 10 | 7 | Traingd | LOGSIG | PURELIN | DIVIDERAND | LERANGDM | 0.077421 |

From the above experiments, this study has found out that MLP ANN as the best model to predict the severity of wheat stem rust using one hidden layer with eight neurons, logsig and purelin as transfer function, learngdm as learning function, trainlim as training function and 70/15/15 data split with MSE value is 0.049962. This proposed MLP model diagram is shown below.

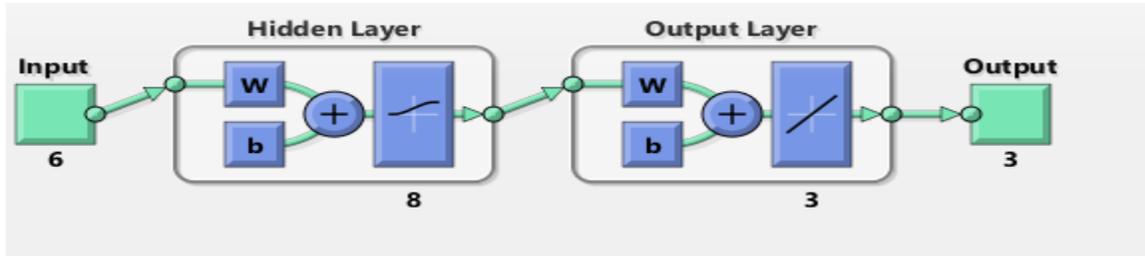

Figure 2.Proposed MLP Model

4.1 Determination of Appropriate Architecture of RBFNN Model

Learning or training a neural network is a process by means of which the network adapts itself to a stimulus by making proper parameter adjustment, resulting in the production of desired response. Hence, in order to achieve the similar approximation or classification accuracy and in addition to the required number of RBFNN units, the following parameters are determined by the training process of RBFNN.

I. The selection of the radial basis centers

This study used the orthogonal least-squares (OLS)method [30] [31] [32] is an efficient way for subset model selection. The approach chooses and adds RBFNN centers one by one until an adequate network is constructed. All the training examples are considered as candidates for the centers, and the one that reduces the MSE the most is selected as a new hidden unit.

II. The number of radial basis(hidden) neurons

For a large data set, as the number of hidden neurons is equal to the number of data points, the mapping function can be very expensive to be computed. Broomhead and Lowe [33] showed that the exact interpolation is not a good strategy because of its poor generalization. Considering this problem, Broomhead and Lowe removed the restriction of the exact interpolation function and designed a two-layer network structure where RBFNNs are used as computation units in the hidden layer. This model gives a smoother fit to the data using a reduced number of basis functions which depends on the complexity of the mapping function. So, in this work the basic principle is: 0 as a neuron started training, by checking the output error to make the network automatically increase neurons, after the training sample looping once, using the training sample which make the network produce have the maximum error as the weight vector to generate a new hidden neuron, then recalculating, checking the error of the new network, repeating this process until it reaches the required error or maximum number of hidden neurons.

### III. The selection of the basis function radius (spread)

The larger spread is, the smoother the function approximation. Too large a spread means a lot of neurons are required to fit a fast-changing function. Too small a spread means many neurons are required to fit a smooth function, and the network might not generalize well. In this study, the radial basis network was assumed to be 0.1 to 2.0 with the step size 0.1 [34].

### 4.2 Modeled and Experiment of Proposed RBFNN Approach

In this experiment, for training RBFNN network with appropriate generalization ability, the target error was first considered zero, the maximum number of neurons in the hidden layer were chosen by trial and error in order to minimize the Mean Squared Error (MSE) in the training and validation sets and the width of the center(spread) assumed to be 0.1 to 2.0 with the step size 0.1 [34].

Table 8. Error analysis of developed RBFNN models for different network

| Model No | Spread | No of hidden neurons | Mean Square Error (MSE) | Epoch |
|---|---|---|---|---|
| 1 | 0.1 | 510 | 0.0101568 | 500 |
| **2** | **0.2** | **656** | **0.0100134** | **650** |
| 3 | 0.3 | 746 | 0.0114062 | 700 |
| 4 | 0.4 | 919 | 0.0104852 | 900 |
| 5 | 0.5 | 1064 | 0.0101949 | 1050 |
| 6 | 0.6 | 1118 | 0.0103292 | 1100 |
| 7 | 0.7 | 1157 | 0.0101277 | 1150 |
| 8 | 0.8 | 1190 | 0.0109623 | 1150 |
| 9 | 0.9 | 1218 | 0.0104709 | 1200 |
| 10 | 1.0 | 1230 | 0.0106513 | 1200 |
| 11 | 1.1 | 1277 | 0.01048 | 1250 |
| 12 | 1.2 | 1361 | 0.0102354 | 1350 |

| 13 | 1.3 | 1604 | 0.0101233 | 1600 |
| 14 | 1.4 | 2000 | 0.0115874 | 2000 |
| 15 | 1.5 | 2000 | 0.0130079 | 2000 |
| 16 | 1.6 | 2000 | 0.0148889 | 2000 |
| 17 | 1.7 | 2000 | 0.0169153 | 2000 |
| 18 | 1.8 | 2000 | 0.0187552 | 2000 |
| 19 | 1.9 | 2000 | 0.0196425 | 2000 |
| 20 | 2.0 | 2000 | 0.0209732 | 2000 |

As in the experiment stated in table 8, the researcher assessed different network structures of RBFNN which were carried out in Matlab environment. So, the second model has a good training and test capability. Therefore, the researcher selected the RBFNN model with 6-656-3 network architecture and the spread is 0.2 and the MSE value is 0.0100134.

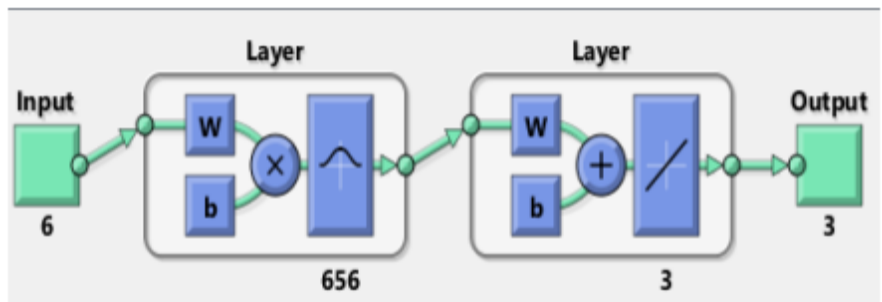

Figure 3. Proposed RBFNN Model

### 4.3 Determination of Appropriate Architecture of GRNN Model

GRNN is the use of the smoothing factor, $\sigma$, which changes the degree of generalization of the network. Therefore, a range of smoothing factors and methods for selecting smoothing factors are tested in the study in order to determine the optimum smoothing factors for model inputs. In this network, there are no training parameters such as the learning rate, momentum, optimum number of neurons in the hidden layer, and learning algorithm as in backpropagation network, but there is a smoothing factor which its optimum value is gained by trial and error. The smoothing factor has to be greater than 0 and can usually range from 0.1 to 1 with good results [35].

#### 4.3.1 Modeled and Experiment of Proposed GRNN Approach

Table 9. Error analysis of developed GRNN models for different Smoothing Factors

| Model No | Smoothing factor | Mean Square Error (MSE) |
|---|---|---|
| **1** | **0.1** | **0.0170** |
| 2 | 0.2 | 0.0443 |
| 3 | 0.3 | 0.0556 |
| 4 | 0.4 | 0.0612 |
| 5 | 0.5 | 0.0655 |
| 6 | 0.6 | 0.0696 |
| 7 | 0.7 | 0.0735 |
| 8 | 0.8 | 0.0771 |
| 9 | 0.9 | 0.0803 |
| 10 | 2.0 | 0.0832 |

According to table 9, which was analyzed based upon on GRNN approach with several value of smooth factors, with the aim to get optimal model. As a result of this, the smooth factor with 0.1 were found to have good fitting accuracy.

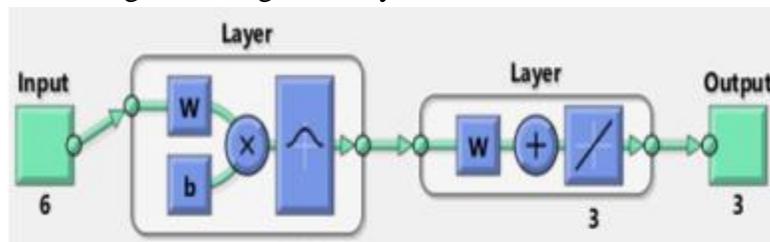

Figure 4. Proposed GRNN Model

5. Model Evaluation and Discussion

In this study, MLP with backpropagation networks with different transfer functions, training functions and learning functions, RBFNN and GRNN were applied to predict the severity of wheat stem rust. So, this section presents each of the following proposed models that are MLP, RBFNN and GRNN. Subsequently, comparative analysis among the three of these models were going to discussed.

    5.1 Wheat Stem Rust Severity Prediction Using MLP with BP

In this study conducted using 10 different training algorithms, namely Trainlm, Trainbfg, Trainrp, Traingdx, Trainscg, Traincgb, Trainoss, Traincgf, Traingdm and Traingd which are used for training MLP with BP. Their characteristics deduced from the experiments are shown table 7.

It was observed that the Traingdm (Gradient descent with momentum backpropagation) had lowest accuracy compared to all other algorithms, whereas trainlm (Levenberg-Marquardt BP) algorithm was capable of achieving very satisfying statistical results with the highest $R^2$ and the lowest mean square error among the other approaches. It recorded the least

MSE(MSE=0.049962) and displayed in Fig.6 revealed overall R value of 0.8629. It also observed to have the least average number of epoch while the entire CPU processing duration.

In terms of speed, traingda (gradient descent BP with adaptive learning rate) had the fastest training speed. Furthermore, although the traingda and traingdx algorithms trained the BP network much faster than the trainlm algorithm. A large learning rate may lead to faster convergence, but it may also cause strong oscillations near the optimal solution or even diverge, while excessively small learning rates result in very long training times. The initiative of adding momentum was to allow the network to respond not only to the local gradient, but also to recent trends in the error surface and allow the network to ignore small features in the error surface. Without momentum, the network can get stuck in a shallow local minimum.

Therefore, the trainlm algorithm was suitable for training the prediction the severity of stem rust MLP ANN model. Similarly, in this work evaluated different transfer function to developed optimal model. Log-sigmoid transfer function (LOGSIG), Hyperbolic tangent transfer function (TANSIG) and Purelin Transfer function are commonly used to MLP.

By considering MSE and correlation coefficient it can be concluded that LOGSIG performed better than TANSIG in the hidden layer and Purelin in the output layer have good performance compare than other transfer function. Linear is also used in the output layer, when you want a real-numbered output (regression). Moreover, the investigator evaluated divide function and learning function. The experiments showed dividerand function have good capability to developed the proposed model. The divide ratios for training, testing and validation are 0.7, 0.15 and 0.15, respectively. In the case of learning function, the researcher evaluated learngdm and learngd. The result shown learngd have better performance than learngdm considering mean square error and coefficient relation.

In general, in the MLP ANN architecture with 8 neurons in the one hidden layer, trainlim as training function, as a transfer function Logsigmoid and Purelin in the hidden and output layer, learngm and divierand used as leraning function and divide function respectively delivered the best results.

Then the network was trained and tested 7 times (7 iterations) and among them, the best network performance based on the mean squared error (MSE) and the correlation coefficient was selected. The results showed that the mean square error of the selected model was 0.049962 and overall coefficient relation was 0.8629. The overall the selected MLP prediction results are shown below. The results of the MLP network model demonstrate good agreement with the measured values.

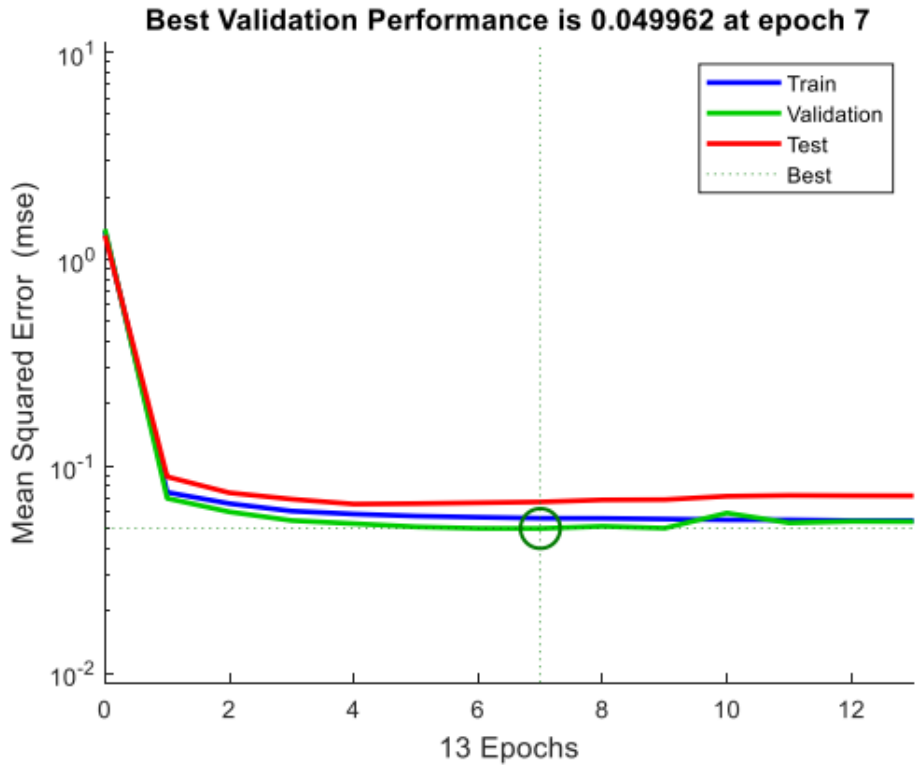

Figure 5. The Performance Graph of ANN Training Process

The fig 5 represents the performance graph. It is displayed when the performance button in Fig5. is clicked. This graph allows the user to know the current status of the training process.

The X – axis indicates the number of iterations (13 Epochs). The Y - axis represents the MSE occurred for each iteration. The line graph plotted in blue color represents the training results. The graph with green color represents the validation results and the graph in red color represents the test results.

The performance graph is computed for every iteration in the training process and the graph in which all the three results of training, validation and testing coincide at almost all points is chosen to be the best performance. At that point of time the training should be stopped and no further iteration should be proceeded. It means that no further training is required and if done, may predict the results wrongly. The fig 5. shown the best performance was occurred in the seventh iteration(epoch) with a 0.049962 MSE value.

This performance plot shows no noticeable problems. The validation and test curves are very similar. So, the validation and the test curves do not indicate overfitting.

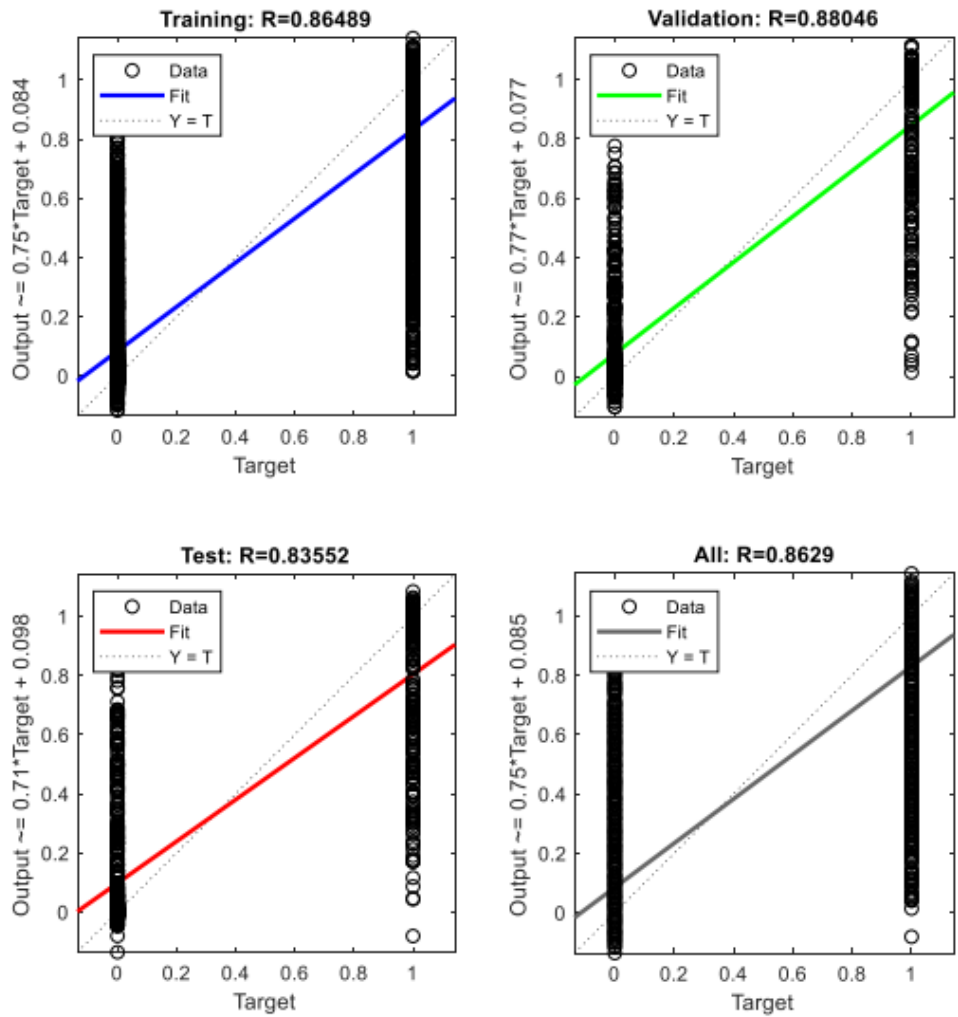

Figure 6. Regression plots of the chosen MLP network

In the fig. 6 illustrated the three plots represent the training, validation, and testing data. The dashed line in each plot represents the perfect result – outputs = targets. The solid line represents the best fit linear regression line between outputs and targets. The R value is an indication of the relationship between the outputs and targets. If R = 1, this indicates that there is an exact linear relationship between outputs and targets. If R is close to zero, then there is no linear relationship between outputs and targets. In this diagram shown the training data indicates a good fit. The validation and test results also show large R values.

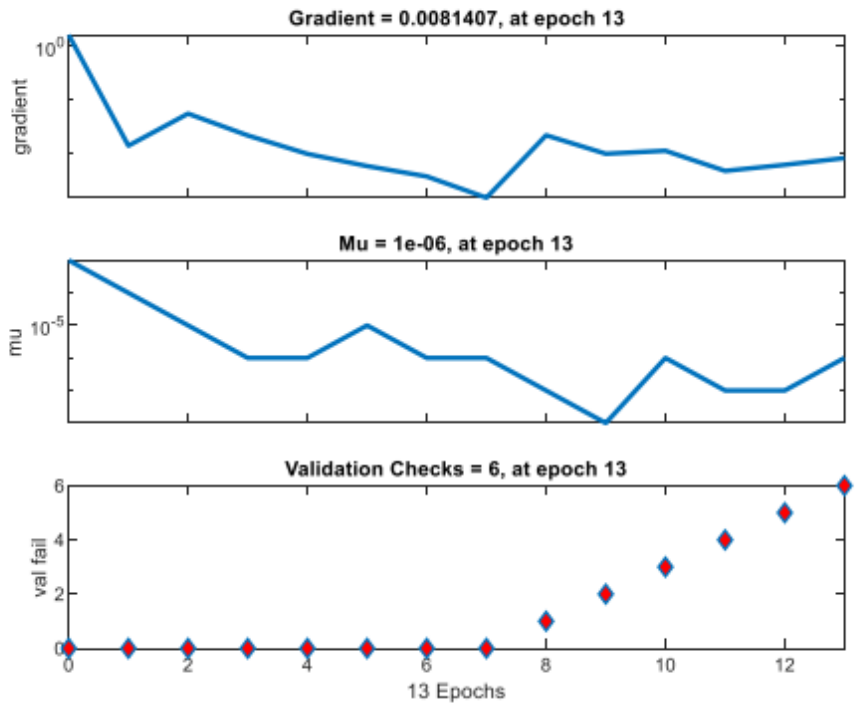

Figure 7. Training state of Chosen MLP network

Figure 7. shows variation in gradient coefficient with respect to number of epochs. The final value of gradient coefficient at epoch number 13 is 0.0081407 which is approximate near to zero. Minimum the value of gradient coefficient better will be training and testing of networks. From figure it can be seen that gradient value goes on decreasing with increase in number of epochs.

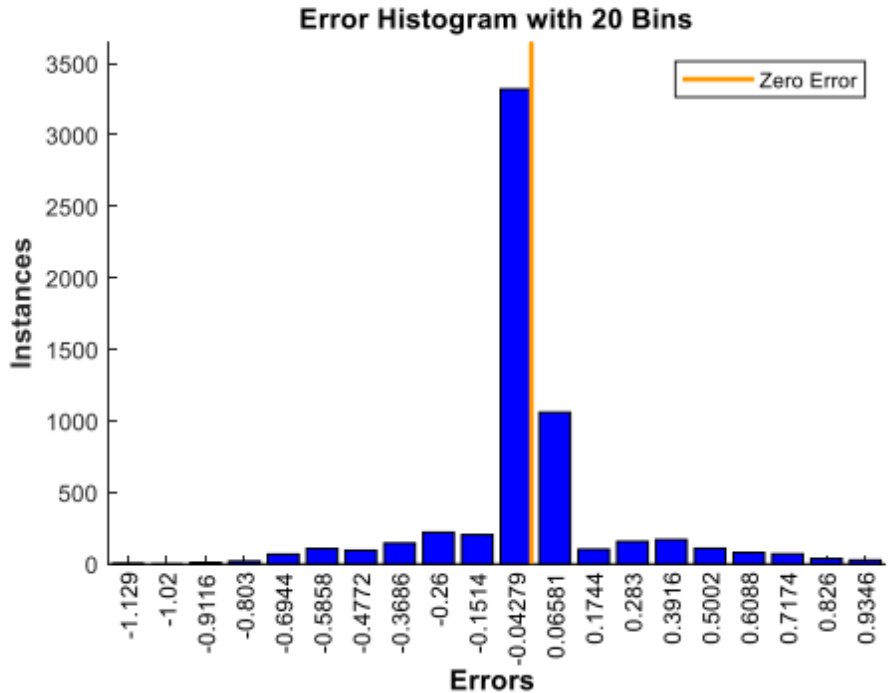

Figure 8. Error Histogram plot of the selected MLP Model

In the above figure shown the total error from neural network ranges from -1.129 (leftmost bin) to 0.93346 (rightmost bin).

## 5.2 Wheat **Stem Severity Prediction using RBFNN**

The researcher tried to designed and trained the RBFNN using different number of radial basis neurons in single hidden layer and spread factor until to get optimum RBFNN model. The experiments have shown the best result are obtained for RBFNN with 656 hidden neurons and the optimal spread factor was 0.2. Based on the experiments the selected RBFNN had a minimum mean square error with 0.0100134 and large R value with 0.97726. The overall performance of selected model shows in below diagrams.

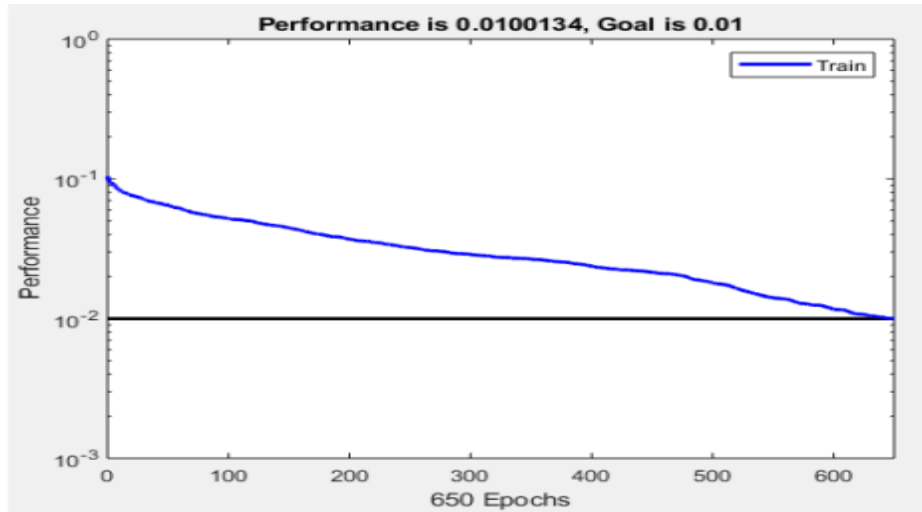

Figure 9. Error variation of radial basis function neural network in the training phase

The above figure shown the error variation of radial basis neural network the selected model and indicated when the iteration increases the mean square error was decrease. So, radial basis networks have good adaptive learning ability and stable prediction ability.

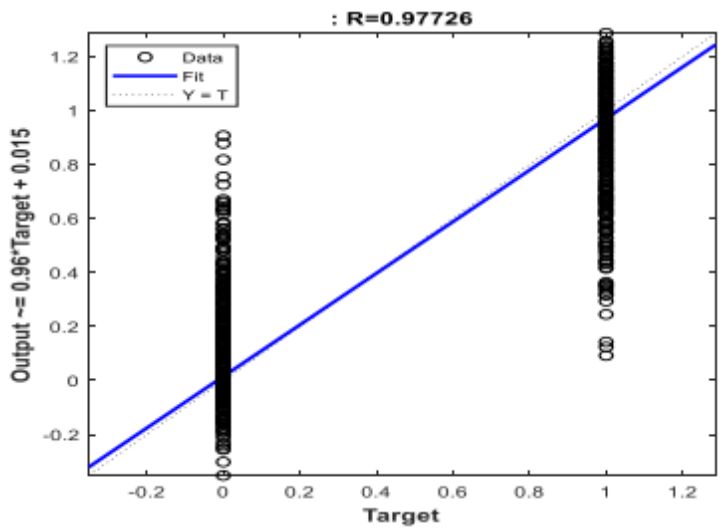

Figure 10 Regression plots of RBFNN Model

In the above figures shown a linear regression analysis between the network outputs and the corresponding targets. Therefore, the diagrams shown the selected RBFNN had a good performance within high R values. i.e. high R-values are indication of good results.

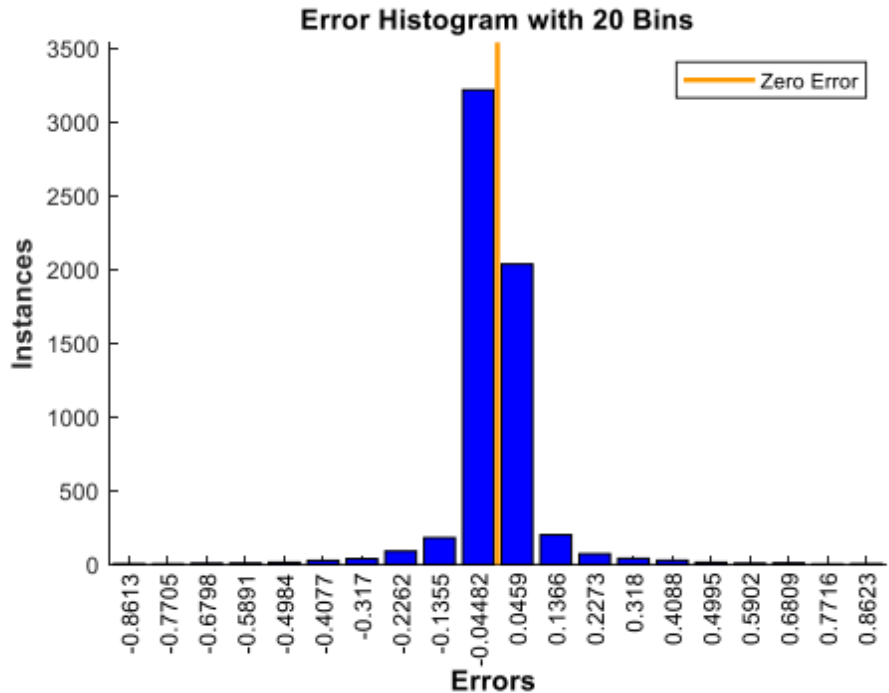

Figure 11. Error Histogram plot of the selected RBFNN Model

Fig 11. Show the error histogram plot for training data for additional verification performance. Bins are the number of vertical bars you are observing on the graph. The total error from neural network ranges from -0.8613 (leftmost bin) to 0.8623 (rightmost bin). This error range is divided into 20 smaller bins. Each vertical bar represents the number of samples from your dataset, which lies in a particular bin. Most of the data falls on zero error line which provides an idea to check the outliers to determine if the data is bad, or if those data points are different than the rest of the data.

### 5.3 Wheat Stem Severity Prediction using GRNN

The GRNN learning method simply stores the training patterns and processes them through a nonlinear smoothing function to determine the component output probability density functions. In this paper evaluated the different values of σ (the smoothing factor) for trained the network, and their corresponding MSE values. The σ values 0.1 can fit data very closely, with R-values was 0.96158 and the mean square error was 0.0170.

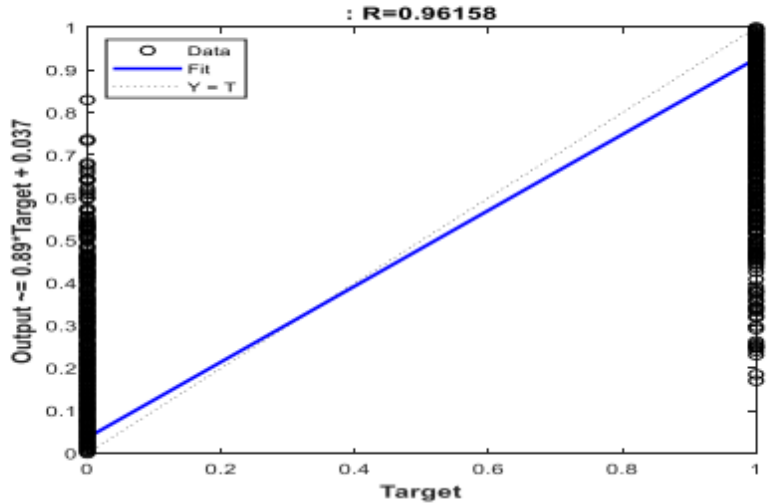

Figure 12. Regression plots of GRNN

The above figures shown how accurately your trained model fits the dataset. The value of R-squared is close to 1 (good) then it shows that the model prediction is very close to the actual dataset.

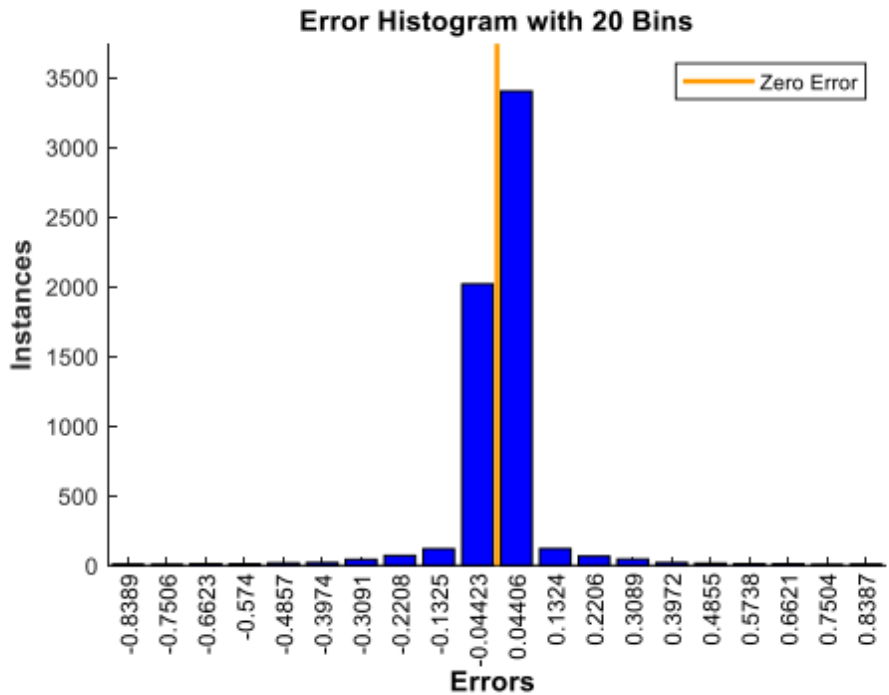

Figure 13. Error histogram plot for the selected GRNN training Data

In the above diagram shown the total error from neural network ranges from -0.8389 (leftmost bin) to 0.83787 (rightmost bin). This error range is divided into 20 smaller bins.

## 5.4 Comparative Analysis Between MLP, GRNN and RBFNN Models

The above experiments have been evaluated all three models and selected best achieved results for the MLP, RBFNN and GRNN ANN models with computed values of R, RMSE, MSE and MAE. Subsequently, compare the three selected ANN models (MLP, RBFNN and GRNN). The obtained results confirm the superiority of the RBFNN technique over the MLP and GRNN technique in most the training cases. Namely, RBFNN network architecture with high Determination coefficient, and low MBE, MAE, MAPE and RMSE values. But, according to the test results GRNN and RBFNN models provide RMSE, MAE, and $R^2$ values close to each other. The GRNN estimations are slightly better than those of the RBFNN. Also, GRNN took less time to train and simple to train compares to MLP and RBFNN. Therefore, GRNN had the optimal ANN model to predict the severity of wheat stem rust. The overall comparison of the three models shown in Table7.

Table 11. Comparison analysis of RBFNN, MLP and GRNN Models

| Model | Training | | | | Testing | | | |
|---|---|---|---|---|---|---|---|---|
| | (RMSE) | (R) | Determination coefficient ($R^{2)}$) | Mean Absolute error (MAE) | RMSE | R | $R^2$ | MAE |
| MLP | 0.224 | 0.86489 | 0.7450 | 0.0520 | 0.07 | 0.835 | 0.70 | 0.036 |
| RBFNN | 0.10 | 0.97726 | 0.955 | 0.0408 | 0.03 | 0.98 | 0.81 | 0.028 |
| GRNN | **0.13** | **0.96158** | **0.925** | **0.0488** | **0.010** | **0.99** | **0.98** | **0.018** |

## 5.5 Discussion
The aim of this study is to develop ANN model which predict the severity of wheat stem rust and that ensures control measures, and it is especially valuable and effective during the application of chemical treatments or biological control agents. In this study, MATLAB tools are used to predict the severity of wheat stem rust in Arsi and Bale zone in Oromia region. Weather data and wheat disease data between 2000 and 2016 are used for training the neural network, while data between 2017and 2018 are used for testing. The investigator used ANN as a prediction tool and chose 6 significant factors in wheat stem rust and gathered data from National meteorological agency, Ethiopia, Sinana Ambo, Kulumusa agricultural institute.

Afterwards, the data served as an input and output variables. These factors namely, mean rainfall, mean maximum, mean minimum temperature, average mean temperature, mean relative humidity and wheat variety used them as input variables and the severity level used as output variable.

In addition, rainfall is the most influential factor. The presented datasets shown some percentage of errors and noises in order to identify the most suitable pattern a cross preprocessing data carried out before the training of the network; both input and output variables were normalized within the range -1 to 1 using a minimax algorithm.

In the MLP approach, the process of building, training, and testing the MLP network models was very complex. Similarly, determining the best results for several network parameters, such as the number of layers and neurons, type of activation functions, training algorithms, learning rates, and momentum, were difficult. The effective way of obtaining good MLP modeling results was to use some trial-and error methods and thoroughly understand the theory of backpropagation.

In RBFNN, there were three parameter that determining the best results. Namely, center and width of a radial basis neural network (RBFNN) and radial basis neurons. There are no local minima problems. The network does not optimize to local minimum solutions because the number of hidden neurons is optimized automatically in the training process. Also, there are no computational time and computer memory problems, especially when there are a large number of input/output training sets, because the network does not have a large number of neurons and weights. The weight values to be optimized exist only on the output side of the hidden layer, while MLP has weights in both sides.

Contrarywise, for the GRNN approach, there was only one parameter (the smoothing factor) that needed to be adjusted experimentally. Moreover, the MLP network performance was very sensitive to randomly assigned initial values. However, this problem was not faced in GRNN techniques. The GRNN approach did not need an iterative training procedure as in the backpropagation method. The local minima problem was also not faced in the GRNN techniques.

6. **Conclusions**

In this study, three types of ANN approach including Multilayer Perceptron (MLP), Radial Basis Neural Network (RBFNN) and GRNN used to develop a model to predicting the severity of wheat stem rust by considering parameters such as mean maximum temperature, mean minimum temperature, mean rainfall, mean average temperature, mean relative humidity and wheat varieties. The presented datasets shown some percentage of errors and noises in order to identify the most suitable pattern a cross preprocessing data carried out before the training of the network; both input and output variables were normalized within the range -1 to 1 using a minimax algorithm. In this study, models were evaluated including MLP-BNN, RBFNN and GRNN by considering of RMSE, MSE, MAPE and $R^2$ to choose the optimal ANN model. The experiment results indicate that

GRNN had effective prediction capability and took less time to train the model compared with other models.

## 7. Future Works

In this study, acceptable model has been developed to predict the severity of wheat stem rust to overcome the problem of unnecessary usage of pesticides. However, some challenges hamper the performance of the study for getting better result, such as the size of data, quality of data and etc. In addition to this, there are many wheat production areas in Ethiopia. This study focused only in two zones in the Oromia regions. The researcher believed that better result would have been attained if many wheat growing areas had been studied under this research. Similarly, in the foreseeable time, provided that different optimization methods are employed, it would be possible to get optimal architecture.